\def\year{2020}\relax
\documentclass[letterpaper]{article} 
\usepackage{aaai20}  
\usepackage{times}  
\usepackage{helvet} 
\usepackage{courier}  
\usepackage[hyphens]{url}  
\usepackage{graphicx} 
\usepackage{graphicx}  
\usepackage{amssymb}
\usepackage{stackengine} 
\usepackage{mathtools}
\usepackage{multirow}
\usepackage{booktabs}

\newcommand{\citet}[1]{\citeauthor{#1}~\shortcite{#1}}

\newcommand{\mytensor}[1]{\ensuremath{\mathcal{#1}}}
\newcommand{\myvector}[1]{\ensuremath{\mathbf{#1}}}
\newcommand{\mymatrix}[1]{\ensuremath{\mathbf{#1}}}

\newcommand{\myrange}[2]{\ensuremath{[{#1}\isep{#2}]}}

\newcommand{\myId}{\ensuremath{\mathbf{Id}}}

\newcommand{\myst}{\ensuremath{^{\text{st}}}\,}
\newcommand{\mynd}{\ensuremath{^{\text{nd}}}\,}
\newcommand{\myrd}{\ensuremath{^{\text{rd}}}\,}
\newcommand{\myth}{\ensuremath{^{\text{th}}}\,}

\newcommand{\myR}{\ensuremath{\mathbb{R}}}

\newcommand{\isep}{\mathrel{{.}\,{.}}\nobreak}

\newcommand{\myT}{^{\top}}

\title{Incremental Multi-domain Learning with Network Latent Tensor Factorization}
\author{Adrian Bulat\thanks{Denotes equal contribution},\textsuperscript{\rm 1} \Large \textbf{Jean Kossaifi}\footnotemark[1],\textsuperscript{\rm 1, 2} \textbf{Georgios Tzimiropoulos,\textsuperscript{\rm 1} Maja Pantic\textsuperscript{\rm 1, 2}}\\ 
\textsuperscript{\rm 1} Samsung AI Center Cambridge \quad \quad
\textsuperscript{\rm 2} Imperial College London\\ 
adrian@adrianbulat.com, jean.kossaifi@gmail.com, georgios.t@samsung.com, maja.pantic@samsung.com 
}
 
\begin{document}

\maketitle

\begin{abstract}
The prominence of deep learning, large amount of annotated data and increasingly powerful hardware made it possible to reach remarkable performance for supervised classification tasks, in many cases saturating the training sets. However the resulting models are specialized to a single very specific task and domain.
Adapting the learned classification to new domains is a hard problem due to at least three reasons: (1) the new domains and the tasks might be drastically different; (2) there might be very limited amount of annotated data on the new domain and (3) full training of a new model for each new task is prohibitive in terms of computation and memory, due to the sheer number of parameters of deep CNNs.
In this paper, we present a method to learn new-domains and tasks incrementally, building on prior knowledge from already learned tasks and without catastrophic forgetting.
We do so by jointly parametrizing weights across layers using low-rank Tucker structure. The core is task agnostic while a set of task specific factors are learnt on each new domain. We show that leveraging tensor structure enables better performance than simply using matrix operations. Joint tensor modelling also naturally leverages correlations across different layers. 
Compared with previous methods which have focused on adapting each layer separately, our approach results in more compact representations for each new task/domain. 
We apply the proposed method to the 10 datasets of the Visual Decathlon Challenge and show that our method offers on average about $7.5\times$ reduction in number of parameters and competitive performance in terms of both classification accuracy and Decathlon score.
\end{abstract}

\section{Introduction} \label{sec:intro}


It is now commonly accepted that supervised learning with deep neural networks can provide satisfactory solutions for a wide range of problems as long as i) the aim is to focus on a single task only, and ii) there is a sufficient availability of labelled training data and computational resources. This is the setting under which Convolutional Neural Networks (CNNs) have been employed in order to provide state-of-the-art solutions for a wide range of Computer Vision problems such as recognition~\cite{krizhevsky2012imagenet,simonyan2014very,he2016deep}, detection~\cite{ren2015faster} and semantic segmentation~\cite{long2015fully,he2017mask} to name a few.

However, visual perception is not just concerned with being able to learn a single task at a time, assuming an abundance of labelled data, memory and computing capacity. A more desirable property is to be able to learn a set of tasks, possibly over multiple different domains, under limited memory and finite computing power. This setting is a very general one and many instances of it have been studied in Computer Vision and Machine Learning under various names.
The main difference comes from whether we vary the \emph{task} to be performed (classification or regression), or the \emph{domain}, which broadly speaking refers to the distribution of the data or the labels for the considered task.

Herein, we are mostly concerned with the problem of multi-domain incremental learning. A key aspect of this setting is that the new task should be learned without harming the classification accuracy and representational power of the original model. 
This is called learning without catastrophic forgetting~\cite{french1999catastrophic,li2017learning}.
Another important aspect is to keep newly introduced memory requirements low: a newly learned model should use as much as possible existing knowledge learned from already learned tasks, i.e. from a practical perspective, it should re-use or adapt the weights of an already trained (on a different task) network.

The aforementioned setting has only recently attracted the attention of the neural network community. Notably, \citet{rebuffi2017learning} introduced the Visual Decathlon Challenge which is concerned with incrementally converting an Imagenet classification model to new ones for another 9 different domain/tasks. 
To our knowledge there are only a few methods that have been proposed recently in order to solve it~\cite{rebuffi2017learning,rebuffi2018efficient,rosenfeld2017incremental,mallya2018piggyback}.
These works all have in common that incremental learning is achieved with layer-specific adapting modules (which are simply called adapters) applied to each CNN layer  separately. Although the adapters have only a small number of parameters, because they are layer specific, the total number of parameters introduced by the adaptation process scales linearly with the the number of layers, and in practice an adaptation network requires about 10\% extra parameters (see also \cite{rebuffi2018efficient}). Our main contribution is to propose a tensor method for multi-domain incremental learning that requires significantly less number of new parameters for each new task. 

\noindent In summary, \textbf{our contributions} are:
\begin{itemize}
    \item 
     We propose to leverage joint parametrization of neural networks for multi-domain learning without catastrophic forgetting. Our method differs from previously proposed layer-wise adaptation methods (and their straightforward layer-wise extensions) by grouping all identically structured blocks of a CNN within a single high-order tensor. 
     \item 
     We perform a thorough evaluation of our model on the 10 datasets of the visual decathlon challenge and show that our method offers on average about $7.5 \times $ reduction in model parameters compared with training a new network from scratch.
     \item
     The joint parametrization of the tensor with a low-rank tensor naturally leverages correlations across different layers. This results in learning more compact representations for each new task/domain.
     \item 
     We show that this tensor approach outperforms methods based on matrix algebra that discard the multi-linear structure in the data.
\end{itemize}

\textbf{Intuitively}, our method first learns, on the source domain, a task agnostic core tensor. This represents a shared, domain-agnostic, latent subspace. For each new domains, this core is specialized by learning a set of task specific factors defining the multi-linear mapping from the shared subspace to the parameter space of each of the domains.

\section{Closely Related Work}
In this section, we review the related work on incremental multi-domain learning and tensor methods.

\textbf{Incremental Multi-Domain Learning} In the context of incremental learning, \citet{rosenfeld2017incremental} and \citet{rebuffi2017learning} introduce the concept of layer adapters. Theses convert each layer\footnote{The last layer typically requires retraining because the number of classes will in general be different.} of a pre-trained CNN (typically on Imagenet) to adapt to a new classification task, for which new training data becomes available. Because the layers of the pre-trained CNN remain fixed, such approaches avoid the problem of catastrophic forgetting \cite{french1999catastrophic,li2017learning} so that performance on the original task is preserved. The method of~\cite{rosenfeld2017incremental} achieves this by computing new weights for each layer as a linear combination of old weights where the combination is learned in an end-to-end manner for all layers via back-propagation on the new task. 
The work in \cite{rebuffi2017learning} achieves the same goal by introducing small residual blocks composed of batch-norm followed by $1\times1$ convolutional layers after each $3\times3$ convolution of the original pre-trained network. Similarly, the newly introduced parameters are learned via back-propagation. The same work introduced the Visual Decathlon Challenge which is concerned with incrementally adapting an Imagenet classification model to \(9\) new and completely different domains and tasks. 
More recently, \cite{rebuffi2018efficient} extends \cite{rebuffi2017learning} by making the adapters to work in parallel with the $3\times3$ convolutional layers. Although the adapters have only a small number of parameters each, they are layer specific, and hence the total number of parameters introduced by the adaptation process grows linearly with the the number of layers. In practice, an adaptation network requires about \(10\%\) extra parameters (see also \cite{rebuffi2018efficient}). In~\cite{mallya2018piggyback} the authors propose to learn to adapt to a new task by learning how to mask individual weights of a pre-trained network. Following the same general idea, in~\cite{morgado2019nettailor}, the authors introduce the so-called NetTailor method. Given an existing pretrained network with generic layers the methods learns how to combine them using a series of small task specific layers.  \cite{guo2019depthwise} makes uses of depthwise separable convolutions in order to build more efficient multi-task networks. In~\cite{mancini2018adding}, the authors propose to learn a task specific binary mask that selects a different set of activations depending of the target task.

Our method significantly differs from these works in that it models groups of identically structured blocks within a CNN with a single high-order tensor. This results in a much more compact representations for each new task/domain, with a latent subspace shared between domains. Only a set of factors, representing a very small fraction of this subspace, needs to be learnt for each new task or domain.

\textbf{Tensor methods} Herein, we focus on methods which have been used to re-parametrize existing deep neural networks. For a review of tensor methods, the reader is referred to existing surveys~\cite{tensor_decomposition_kolda,tensor_decomposition_sidiropoulos,tensor_mining_papalexakis}. 
In deep learning, tensor methods are typically used to speed up computation or to reduce the number of parameters. Convolutional kernels in particular can be decomposed and reformulated more efficiently using CP~\citet{lebedev2015speeding,astrid2017cp} or Tucker~\citet{yong2016compression} decomposition. 
\citet{yang2017deep} proposed a tensor factorization approach to multi-task learning. The weights of several networks (one per task) are parametrized, at each layer, with a low-rank tensor which allows to learn the sharing.
The method of \cite{yunpeng2017sharing} proposed a method to share parameters within a ResNeXt~\cite{xie2017aggregated} block, by applying a Generalized Block Decomposition to a 4-th order tensor. As exemplified here, tensor algebraic operations have the potential to improve deep models. However, a straightforward extension of existing multi-domain adaptation methods to use tensor methods (e.g. \cite{rosenfeld2017incremental}) can result in an adaptation model with a large number of parameters. To improve this, following~\cite{TNet}, we propose to model groups of identically structured blocks within a CNN with a single high-order Tucker tensor, with a task agnostic core and task specific factors. We describe our method in details in the next section.

\begin{figure*}[!htbp]
    \centering
    \includegraphics[width=1\linewidth]{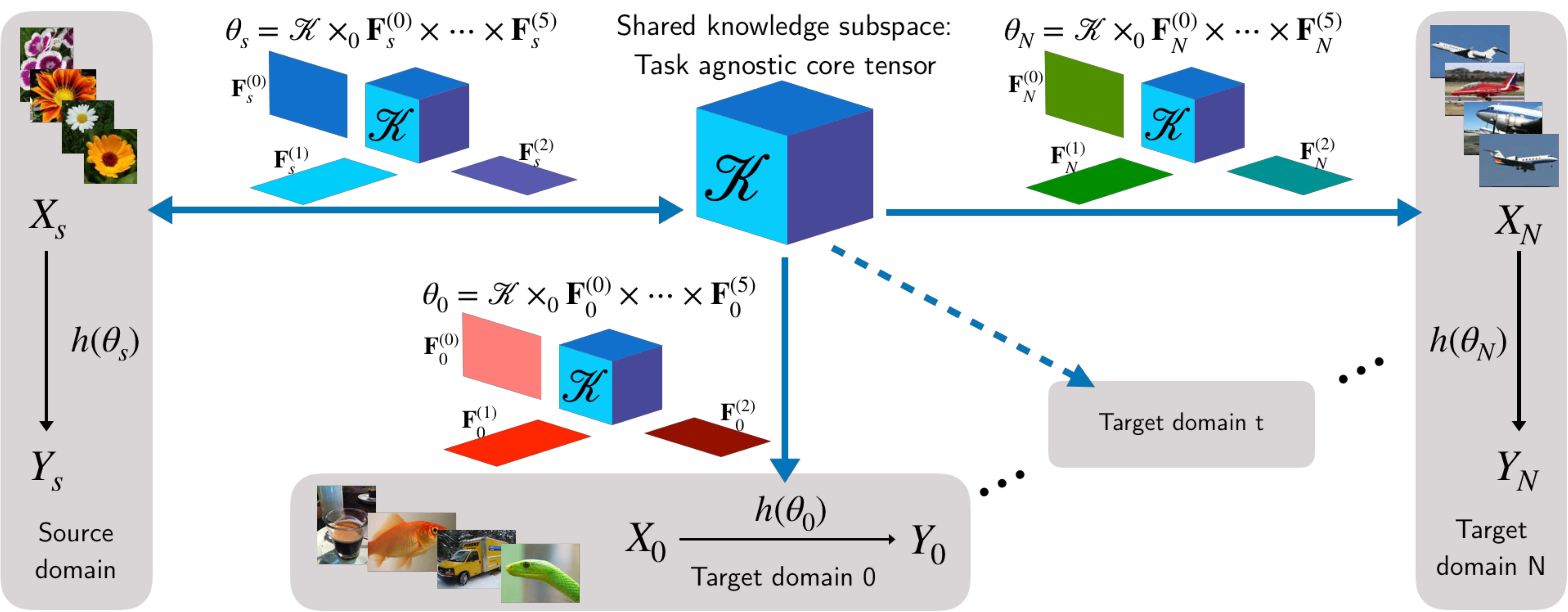}
    \caption{\textbf{Overview of our method} First, a task agnostic core \(\mytensor{K}\) is learned jointly with the domain specific factors \(\mymatrix{F}_s^{(0)}, \cdots, \mymatrix{F}_s^{(5)} \) on the source domain/task (left). For a new target domain/task, the same core is specialized for the new task by training a new set of factors \(\mymatrix{F}_t^{(0)}, \cdots, \mymatrix{F}_t^{(5)} \) (bottom), similarly for any new task \(k \in \myrange{1}{N}\) (right). Intuitively, the core represents a task agnostic subspace, while the task specific factors define the multi-linear mapping from that subspace to the parameter space of each of the domains. Note that here, we represent the \(6\myth\) order tensors in 3D for clarity.}
    \label{fig:method}
\end{figure*}

\section{Method}\label{sec:method}

In this section, we introduce our method (depicted in Figure~\ref{fig:method}) for incremental multi-domain learning, starting by the notation used~(Section.~\ref{ssec:notation}).
By considering a source domain \({X}^s\) and output space \(Y^s\), we aim to learn a function \(h\) (here, a ResNet based architecture)  parametrized by a tensor \(\mytensor{\theta}^s\), \(h( \theta^s) \colon X^s \rightarrow Y^s\). The model and its tensor parametrization are introduced in detail in Section~\ref{ssec:network-architecture}.
The main idea is to then learn a task agnostic latent manifold \(\mytensor{K}\) on the source domain. The parameter tensor \(\mytensor{\theta}^s\) is obtained from \(\mytensor{K}\) with task specific factors \(\mymatrix{F}^{(0)}_s, \cdots, \mymatrix{F}^{(N)}_s\). Given a new target task, we then adapt  \(h\) and learn a new parameter tensor \(\mytensor{\theta}^t\) by specialising \(\mytensor{K}\) with a new set of task specific factors \((\mymatrix{F}^{(0)}_t, \cdots, \mymatrix{F}^{(N)}_t)\). This learning process is detailed in Section~\ref{ssec:multi-domain}.
In practice, most of the parameters are shared in \(\mytensor{K}\), while the factors only contain a fraction of the parameters, which leads to large parameters savings. We offer an in-depth analysis of these space savings in Section~\ref{ssec:complexity-analysis}. 

\subsection{Notation}\label{ssec:notation}
In this paper, we denote \textbf{vectors} (1\myst order tensors) as \(\myvector{v}\), \textbf{matrices} (2\mynd order tensors) as \(\mymatrix{M}\) and \textbf{tensors}, which generalize the concept of matrices for orders (number of dimensions) higher than 2, as \(\mytensor{X}\). \(\myId\) is the identity matrix.
\textbf{Tensor contraction} with a matrix, also called n--mode product, is defined, for a tensor \(\mytensor{X} \in \myR^{D_0 \times D_1 \times \cdots \times D_N}\) and a matrix \( \mymatrix{M} \in \myR^{R \times D_n} \), as the tensor \(\mytensor{T} = \mytensor{X} \times_n \mymatrix{M} \in \myR^{D_0 \times \cdots \times D_{n-1} \times R \times D_{n+1} \times \cdots \times D_N} \), with:
\[
\mytensor{T}_{i_0, i_1, \cdots, i_N} = \sum_{k=0}^{D_n} \mymatrix{M}_{i_n, k} \mytensor{X}_{i_0, \cdots, i_{n-1}, k, i_{n+1}, \cdots, i_N}.
\]

\subsection{Latent Network Parametrization}\label{ssec:network-architecture}
We propose to group all the parameters of a neural network into a set of high-order hyper-parameter tensors~\cite{TNet}.
We do so by collecting all the weights of the neural network into \(3\) parameter tensors \(\mytensor{\theta}^{(0)}\) and \(\mytensor{\theta}^{(2)}\), all of order \(6\).
While the proposed method is not architecture specific, to allow for a fair comparison in terms of overall representation power, we follow~\cite{rebuffi2017learning,rebuffi2018efficient,rosenfeld2017incremental} and use a modified ResNet-26~\cite{he2016deep}. 
The network consists of 3 macro-modules, each consisting of 4 basic residual blocks~\cite{he2016deep}. Each of these blocks contain two convolutional layers with $3\times3$ filters. Following~\cite{rebuffi2017learning}, the macro-modules output 64, 128, and 256 channels respectively. Throughout the network the resolution is dropped multiple times. First, at the beginning of each macro-module using a convolutional layer with a stride of 2. A final drop in resolution is done at the end of the network, before the classification layer, using an adaptive average pooling layer that reduces the spatial dimensions to resolution of $1\times1$ px.

In order to facilitate the proposed grouped tensorization process, we moved the feature projection layer (a convolutional layer with $1\times1$ filters), required each time the number of features changes between blocks, outside of the macro-modules (i.e. we place a convolutional layer with a $1\times1$ kernel before the $2\mynd$ and $3\myrd$ macro-modules). 

We closely align our tensor re-parametrization to the network structure by grouping together all the convolutional layers within the same macro-module. For each macro-module $b\in\{0,1,2\}$, we construct a $6\myth$-order tensor collecting the weights in that group:
\begin{equation}
\mytensor{\theta}^{(b)} \in  \myR^{D_0 \times D_1 \times \cdots \times D_5}
\end{equation}
where $\mytensor{W}^{b}$ is the tensor for the $b$\myth macro-module. The 6 dimensions of the tensor are obtained as follows: $D_0 \times D_1 \times D_2 \times D_3$ corresponds to the shape of the weights of a particular convolution layer and represents the number of output channels, number of input channels, kernel width and kernel height respectively. The $D_4$\myth mode corresponds to the number of basic blocks per residual module (2 in this case) and, $D_5$ corresponds to the number of residual blocks present in each macro-module (4 for the specific architecture used). 

Our model should be compared with previous methods for incremental multi-domain adaptation like \cite{rosenfeld2017incremental} (the method of \cite{rebuffi2017learning} can be expressed in a similar way) which learn a linear transformation per layer. In particular, \cite{rosenfeld2017incremental} learns a 2D adaptation matrix $\mymatrix{F} \in \myR^{D_0 \times (D_1\times D_2 \times D_3)}$ per convolutional layer. Moreover, prior work on tensors (e.g. \cite{yong2016compression}) has focused on standard layer-wise modelling with a $4^{\myth}$-order convolutional filter, of shape $D_0 \times D_1 \times D_2 \times D_3$. In contrast, our model has two additional dimensions incorporating intra-architecture correlations and can accommodate an arbitrary number of dimensions depending on the architecture used. Our approach is in particular not architecture specific and can be used for various network architectures.

\subsection{Multi-Domain Tensorized Learning}\label{ssec:multi-domain}
We now consider a scenario with \(T\) tasks, from potentially very different domains. 
The traditional approach would consist in learning as many models, one for each task. In our framework, this would be equivalent to learning one parameter tensor \(\mytensor{\theta}^{(b)}_d\) independently for each task \(d\) and macro-module \(b\). Since the reasoning is the same for each of the $3$ macro-modules, for clarity and without loss of generality, we omit the \(b\) in the remaining of the paper.
We propose that all the parameters are obtained from a shared latent subspace, modelled by a task agnostic tensor \(\mytensor{K}\). 
The (multi-linear) mapping between this task agnostic core and the parameter tensor is then given by a set of task specific factors \((\mymatrix{F}_{s}^{(0)}, \cdots \mymatrix{F}_{s}^{(5)})\) that specialize the task agnostic subspace for the source domain \(s\). Specifically, we write, for the source domain \(s\):
\begin{equation}
    \mytensor{\theta}_{s} = \mytensor{K} \times_0 \mymatrix{F}_{s}^{(0)} \times_1 \mymatrix{F}_{s}^{(1)} \times \cdots \times_5 \mymatrix{F}_{s}^{(5)}, \label{eq:core}
\end{equation}
where $\mytensor{K} \in \myR^{D_0 \cdots \times D_5}$ is a \textbf{task-agnostic} full rank core \textbf{shared} between all domains and $(\mymatrix{F}^{(0)}_s,\mymatrix{F}^{(1)}_s,\cdots,\mymatrix{F}^{(5)}_s)$ a set of \textbf{task specific} projection factors (for domain \(s\)). We assume here that the task used to train the shared core is a general one with many classes and large amount of training data (here, Imagenet classification). Moreover, a key observation to make at this point is that the number of parameters for the factors is orders of magnitudes smaller than the number of parameters of the core.

For each new target domain \(t\), we form a new parameter tensor \(\mytensor{\theta}_t\) obtained from the \emph{same} latent subspace \(\mytensor{K}\). This is done by learning a new set of factors \(( \mymatrix{F_{t}}^{(0)}, \cdots \mymatrix{F_{t}}^{(5)})\) to specialize \(\mytensor{K}\) for the new task:
\begin{equation}
    \mytensor{\theta}_t = \mytensor{K} \times_0 \mymatrix{F_{t}}^{(0)} \times_1 \mymatrix{F_{t}}^{(1)} \times \cdots \times_5 \mymatrix{F_{t}}^{(5)} \label{eq:new_task}
\end{equation}
Note that the new factors represent only \emph{a small fraction} of the total number of parameters, the majority of which are contained within the shared latent subspace.
By expressing the new weight tensor $\mytensor{\theta}_{t}$ as a function of the factors $\mymatrix{F_{t}}$, one can learn them on the new task given that labelled data are available in an end-to-end manner via back-propagation. This allows to efficiently adapt the domain agnostic subspace to the new domains while retaining the performance on the original task, and training only a small number of additional parameters. Fig.~\ref{fig:method} shows a graphical representation of our method, where the weight tensors have been simplified to 3D for clarity.

\paragraph{Auxiliary loss function:}
To prevent degenerate solutions and facilitate learning, we additionally explore orthogonality constraints on the task specific factors. This type of constraints have been shown to encourage regularization, improving the overall convergence stability and final accuracy~\cite{brock2016neural,bansal2018can}.
In addition, by adding such constraint, we aim to enforce the factors of the decomposition to be full-column rank, which would ensure that the core of the decomposition preserves essential properties of the full weight tensor such as the Kruskal rank~\cite{jiang2017tensor}.
In practice, rather than a hard constraint, we add a loss to the objective function: 
\begin{equation}\label{eq:orthogonality}
    \mathcal{L} =  \lambda \sum_{k=0}^5 \| \left(\mymatrix{F_{k}}^{(k)}\right)\myT \mymatrix{F_{k}}^{(k)} - \myId\|_F^2.
\end{equation}
The regularization parameter \(\lambda\) was validated on a small validation set.

\subsection{Complexity Analysis}\label{ssec:complexity-analysis}

In terms of unique, task specific parameters learned, our grouping strategy is significantly more efficient than a layer-wise parametrization. For a given group of convolutional layers, in this work defined by the macro-module structure present in a ResNet architecture, we can express the total number of parameters for a Layer-wise Tucker case (this is not proposed in this work but mentioned here for comparison purposes) as follows:
\(
    N_{\text{layerwise}} = (D_4 \times D_5) \times (\sum_{k=0}^3 D_n R_k).
    \label{eq:parameters-layer-wise-generic}
\)

In particular, in the case of a full rank decomposition, by denoting \(L = D_4 \times D_5\) the number of convolutional layers, we get:
\begin{equation}
    N_{\text{layerwise}} = \underbrace{(D_4 \times D_5)}_L \times (D_0^2 + D_1^2 + D_2^2 + D_3^2),
    \label{eq:parameters-layer-wise}
\end{equation}
where $L$ is the number of re-parametrized layers in a given group.

For the linear case~\cite{rosenfeld2017incremental}, we have that $D_0=D_1=D_c$, and the number of parameters simplifies to:
\(
    N_{\text{linear}} = \underbrace{(D_4 \times D_5)}_L \times D_c^2
\)
As opposed to this, for our proposed method, by grouping the parameters together into a single high-order tensor, the total number of parameters is:
\(
    N_{\text{T-Net}} = \sum_{k=0}^5 D_n R_k.
\)
For the full-rank case \( (D_n = R_k) \), this simplies to:
\begin{equation}\label{eq:parameters-grouped}
    N_{\text{T-Net}} = D_0^2 + D_1^2 + D_2^2 + D_3^2 +
            \underbrace{D_4^2 + D_5^2}_{(\frac{L}{D_5})^2 + (\frac{L}{D_4})^2}
\end{equation}
Note that here, \(D_4 = 2\) and \(D_5 = 4\) so \(D_4^2 + D_5^2 \leq \frac{L^2}{4}\).

Because in practice $ (D_0^2 + D_1^2 + D_2^2 + D_3^2) \gg L^2 $, by using the proposed method, we achieve
\(
    \frac{N_{\textrm{layerwise}}}{N_{\text{T-Net}}} \approx L
\)
times less task-specific parameters. 

Substituting the variables from Eq.~(\ref{eq:parameters-layer-wise}) and Eq.~(\ref{eq:parameters-grouped}) with the numerical values specific to the architecture used in this work we obtain in total: $ N_{\text{layerwise}} =1,376,688$ 
parameters. By contrast, using the same setting for our proposed method, we get $N_{\text{T-Net}} =172,068$, thus verifying $\frac{N_{\text{layerwise}}}{N_{\text{T-Net}}} \approx 8 = L$.

Making the same assumptions as for the linear case, given that we use square convolutional kernels (i.e. $D_2=D_3=D_n$), and $D_c \gg D_n$, Eq.~(\ref{eq:parameters-grouped}) becomes: $N_{\text{T-Net}} \leq 2 D_c^2 + \frac{L^2}{4}$, resulting in $\approx \frac{L}{2}$ less parameters than in the linear case ($\frac{L}{2}=4$ for the model used).

\paragraph{Conclusion:} Our proposed approach uses \textbf{$L$ times less parameters per group} than the layers-wise Tucker decomposition and \textbf{$\frac{L}{2}$ times less parameters than the layer-wise linear decomposition}. For the ResNet-26 architecture used in this work $L=8$.

\begin{table*}[tp]
	\centering
	\setlength{\tabcolsep}{4pt}
    \resizebox{1\linewidth}{!}{%
        \begin{tabular}{ l c c c c c c c c c c c c c c }
		    \toprule
		    \multicolumn{2}{c}{} & \multicolumn{10}{c}{\textbf{Dataset}} & \multicolumn{1}{c}{} \\
			\cmidrule{3-12}
			\multicolumn{1}{c}{\multirow{-2.5}{*}{\textbf{Model}}}
			   &  \multicolumn{1}{c}{\multirow{-2.5}{*}{\textbf{\#param}}}
			   &  \textbf{ImNet} &  \textbf{Airc.} &  \textbf{C100} &  \textbf{DPed} &  \textbf{DTD} &  \textbf{GTSR} &  \textbf{Flwr} &  \textbf{OGlt} &  \textbf{SVHN} &  \textbf{UCF} 
			   & \multicolumn{1}{c}{\multirow{-2.5}{*}{\textbf{Mean}}}
			   & \multicolumn{1}{c}{\multirow{-2.5}{*}{\textbf{EScore}}}
			   & \multicolumn{1}{c}{\multirow{-2.5}{*}{\textbf{Score}}} \\
			\midrule
			\multicolumn{1}{c}{Model \textbackslash Number of images} & - & 1.3M & 7K & 50K & 30K & 4K & 40K & 2K & 26K & 70K & 9K & - & - & - \\
			\midrule
		    Rebuffi et al. 2017 & 2$\times$ & 59.23  & 63.73  & 81.31  & 93.30  & 57.02  & 97.47  & 83.43  & 89.82 & 96.17 & 50.28 & 77.17 & 1322 &  2643\\
			Rosenfeld et al. 2017 & $2\times$  & 57.74 & 64.11  & 80.07  & 91.29  & 56.54  & 98.46  & 86.05  & 89.67  & 96.77 & 49.38 & 77.01 & 1425 & 2851\\
			Mallya et al. 2018 & $1.28\times$ & 57.69 & 65.29 & 79.87 & 96.99 & 57.45 & 97.27 & 79.09 & 87.63 & 97.24 & 47.48 & 76.60  & 2217 & 2838   \\
			S. Adap.~(Rebuffi et al. 2017) & $2\times$ & 60.32  & 61.87  & 81.22  & 93.88  & 57.13 & 99.27  & 81.67  & 89.62 & 96.57 & 50.12 & 77.17  & 1580 & 3159 \\
			P. Adap.~(Rebuffi et al. 2017) & $2\times$ & 60.32  & 64.21  & 81.91  & 94.73  & 58.83  & 99.38  & 84.68  & 89.21 & 96.54 & 50.94 & 78.07  & 1706 & 3412 \\
			P. SVD~(Rebuffi et al. 2017) & $1.5\times$ & 60.32  & 66.04  & 81.86  & 94.23  & 57.82  & 99.24  & 85.74  & 89.25 & 96.62 & 52.50 & 78.36  & 2265 & 3398 \\
			NetTailor~(Morgado et al. 2019) & $1.5\times$ & 61.42  & 75.07  & 81.84  & 94.68  & 61.28  & 99.52  & 88.53  & 90.09 & 96.44 & 49.54 & 79.64  & 2496 & 3744 \\
			SpotTune~\cite{guo2019spottune} & $11\times$ & 60.32  & 63.91  & 80.48  & 94.49  & 57.13  & 99.52  & 85.22  & 88.84 & 96.72 & 52.34 & 77.89  & 328 & 3612 \\
			Depthwise~\cite{guo2019depthwise}* & $1\times$ & 63.99  & 61.06  & 81.20  & 97.00  & 55.48  & 99.27  & 85.67  & 89.12 & 96.16 & 49.33 & 77.82 & 3507 & 3507 \\
			Bin. mask~\cite{mancini2018adding} & $1.29\times$ & 60.8  & 52.8  & 82.0  & 96.2  & 58.7  & 99.2  & 88.2  & 89.2 & 96.8 & 48.6 & 77.2  & - & - \\
			$BA^2$~\cite{berriel2019budget} & $1.03\times$ & 56.9  & 49.9  & 78.1  & 95.5  & 55.1  & 99.4  & 86.1  & 88.7 & 96.9 & 50.2 & 75.7  & 3105 & 3199 \\
			\midrule 
			\textbf{Ours} & $1.35\times$ & 61.48  & 67.36  & 80.84  & 93.22  & 59.10  & 99.64 & 88.99  & 88.91 & 96.95 & 47.90 & 78.43  & 2656 & 3585  \\
			\bottomrule 
		\end{tabular}
	}
	\caption{\textbf{Comparison to the state-of-the-art:} Top-1 classification accuracy (\%) and overall decathlon scores on all \(10\) dataset from the Visual Decathlon challenge. Our method is generic, applicable to any network architecture. It offers a good balance between accuracy and number of task-specific parameters used.}
	\label{tab:main-results}
\end{table*}

\section{Experimental Setting}
In this section, we detail the experimental setting, metrics used and implementation details.

\paragraph{Datasets:}
We evaluate our method on the \(10\) different datasets from very different visual domains that compose the Decathlon challenge~\cite{rebuffi2017learning}. Note that this dataset where modified in~\cite{rebuffi2017learning}, mainly by resizing and cropping them to the same resolution ($72\times72$px). This challenge assesses explicitly methods designed to solve incremental multi-domain learning without catastrophic forgetting.
\textbf{Imagenet}~\cite{russakovsky2015imagenet} contains \(1.2\) millions images distributed across \(1000\) classes. Following~\cite{rebuffi2017learning,rebuffi2018efficient,rosenfeld2017incremental}, this was used as the source domain to train the shared low-rank manifold for our model as detailed in Eq.~(\ref{eq:core}). 
The \textbf{FGVC-Aircraft Benchmark} (Airc.)~\cite{maji2013fine} contains 10,000 aircraft images across 100 different classes; \textbf{CIFAR100} (C100)~\cite{krizhevsky2009learning} is composed of \(60000\) small images in \(100\) classes; \textbf{Daimler Mono Pedestrian Classification Benchmark} (DPed) \cite{munder2006experimental} is a dataset for pedestrian detection (binary classification) composed of 50,000 images; \textbf{Describable Texture Dataset} (DTD)~\cite{cimpoi2014describing} contains \(5640\) images, for \(47\) texture categories; the \textbf{German Traffic Sign Recognition} (GTSR) Benchmark \cite{stallkamp2012man} is a dataset of \(50,000\) images of  \(43\) traffic sign categories; \textbf{Flowers102} (Flwr)~\cite{nilsback2008automated} contains \(102\) flower categories with between \(40\) and \(258\) images per class; \textbf{Omniglot} (OGlt) \cite{lake2015human} is a dataset of \(32000\) images representing \(1623\) handwritten characters from \(50\) different alphabets; the \textbf{Street View House Numbers} (SVHN)~\cite{netzer2011reading} is a digit recognition dataset containing \(70000\) images in \(10\) classes. Finally, \textbf{UCF101} (UCF)~\cite{soomro2012ucf101} is an action recognition dataset composed of 13,320 images representing 101 action classes. 

\paragraph{Metrics:} 
We follow the evaluation protocol of the Decathlon Challenge and report results in terms of mean accuracy and decathlon score S, computed as follows:
\begin{equation}
    S = \sum_{t=1}^{N} \beta_{t}\max\{0, E_{t}^{reference}-E_{t}\}^{\lambda_{t}},
    \label{eq:decathlon-score}
\end{equation}
where $E_{t}^{reference}$ is considered to be the upper limit allowed for a given task $t$ in order to receive points, $\lambda_{t}$ is an exponent that controls the reward proportionality, and $\beta_{t}$ a scalar that enforces the limit of 1000 points per task. $E^{reference} = 2E^{baseline}$ where $E_{baseline}$ is the strong baseline from~\cite{rebuffi2017learning}. 

One key limitation of this metric is that it doesn't take in consideration the model capacity or the methods compression abilities. As such, following~\cite{mancini2018adding,berriel2019budget} we also report an efficiency based scored (\emph{EScore}) that is simply computed by dividing the decathlon score by the relative number of parameters required across all tasks with respect to the original ResNet-26 network from~\cite{rebuffi2017learning}.

\paragraph{Implementation details:}
We first train our adapted ResNet-26 model on ImageNet for 90 epochs using SGD with momentum ($0.9$), using a learning rate of $0.1$ that is decreased in steps by $10\times$ every 30 epochs. To avoid over-fitting, we use a weight decay equal to $10^{-5}$. During training, we follow the best practices and randomly apply scale jittering, random cropping and flipping. We initialize our weights from a normal distribution $\mathcal{N}(0, 0.002)$, before decomposing them using Tucker decomposition (Section \ref{sec:method}). Finally, we train the obtained core and factors (via back-propagation) by reconstructing the weights on the fly.

For the remaining \(9\) domains, we load the task-independent core and the factors trained on imagenet, freeze the core weights and only fine-tune the factors, batch-norm layers and the two $1\times1$ projection layers, all of which account for $\approx 3.5\%$ of the total number of parameters in total. The linear layer at the end of the network is trained from scratch for each task and was initialized from a uniform distribution. Depending on the size of the dataset, we adjust the weight decay to avoid overfitting ($10^{-5}$ for larger datasets) and up to $0.005$ for the smaller ones (e.g. Flowers102). 

We used PyTorch~\cite{pytorch} to implement and train the models and TensorLy~\cite{tensorly} for all tensor operations.

\section{Results}\label{sec:results}
First, we assess the performance of the proposed approach~(Section~\ref{ssec:results-sota}) and compare it with the state-of-the-art on the challenging Visual Decathlon~\cite{rebuffi2017learning}. In Section~\ref{ssec:results-transfer-learning}, we study of the method, including the importance of source dataset, the influence of the amount of data on the overall performance and the effect of the constraints imposed on the core and factors of the model.

\subsection{Comparison with State-of-the-Art}\label{ssec:results-sota}
Herein, we compare against the current state-of-the-art methods on multi-domain transfer learning~\cite{rebuffi2017learning,rebuffi2018efficient,rosenfeld2017incremental,mallya2018piggyback} on the decathlon dataset.
We train our core subspace on ImageNet and incrementally adapt to all \(9\) other domains.
We report, for all methods, the relative increase in number of parameters (per domain), the top-1 accuracy on each of the \(10\) domain, as well as the average accuracy and overall challenge score, Table~\ref{tab:main-results}.

Overall, our approach offers competitive results in terms of both accuracy and efficiency (i.e. number of introduced task-specific parameters), offering  a good balance between the two.
When compared within the same class of methods, that directly apply a form of matrix decomposition~\cite{rebuffi2018efficient,rosenfeld2017incremental} our approach outperforms all of them, including the joint compression method of~\cite{rebuffi2018efficient} (denoted as ``Parallel SVD'') that takes advantage of the data redundancy in-between tasks. We could go further and impose a weight sharing within the prediction layers by removing the flattening and fully-connected layers altogether, replace them with a tensor regression layer (TRL)~\cite{kossaifi2018tensor}. Our approach can then be readily applied to the low-rank Tucker tensor of the TRL. Furthermore, our method could be combined with binary masking approaches~\cite{berriel2019budget,mallya2018piggyback} that reduce the number of additional parameters by compacting the binary values using bit packing.

\subsection{Inter-class Transfer Learning}\label{ssec:results-transfer-learning}
Most of the recent work on multi-domain incremental learning attempts to transfer the knowledge from a model, pre-trained on a large scale dataset such as ImageNet to another easier dataset and/or task. In this work, we go on step further and explore the efficiency of our transfer learning approach when such source dataset or computational resources are not available, by starting from a model pre-trained on a much smaller dataset. Table~\ref{tab:results-from-cifar100} shows the results for a network pre-trained of CIFAR100. Notice that on some datasets (i.e. GTSRB, OGlt) such model can match or marginally surpass the performance of its Imagenet counterpart. On the other hand, on some of the more challenging datasets (i.e. DTD, aircraft) there is still a large gap. This suggest that the features learned by Cifar-trained model are less generic and diverse. This is due to both the low quantity of available samples and the easiness/overfitting on the original dataset. A potential solution for this may be to enforce a diversity loss, and fine-tune the core jointly on all tasks. However we leave the exploration of this area for future works.
\begin{figure*}[!htbp]
     \begin{tabular}
        {
            @{\hspace{0mm}}c@{\hspace{0mm}} @{\hspace{0mm}}c@{\hspace{0mm}} @{\hspace{0mm}}c@{\hspace{0mm}} @{\hspace{0mm}}c@{\hspace{0mm}}
        }
        \includegraphics[width=0.25\linewidth]{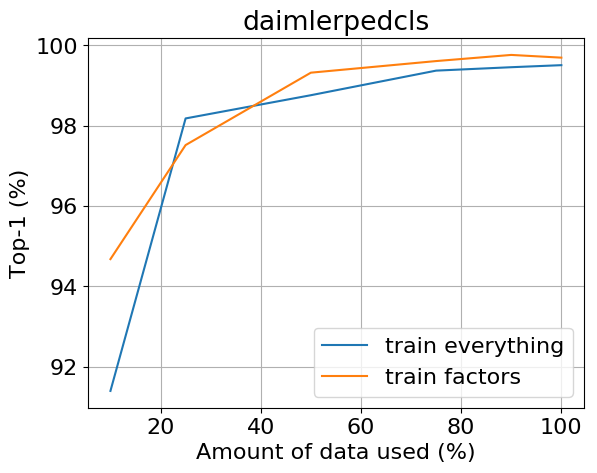} &
        \includegraphics[width=0.25\linewidth]{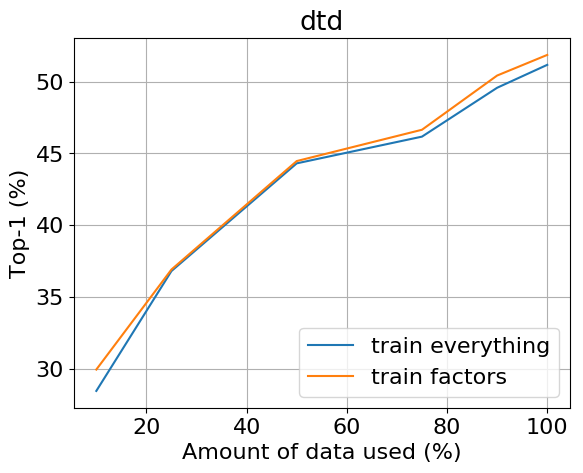} &
        \includegraphics[width=0.25\linewidth]{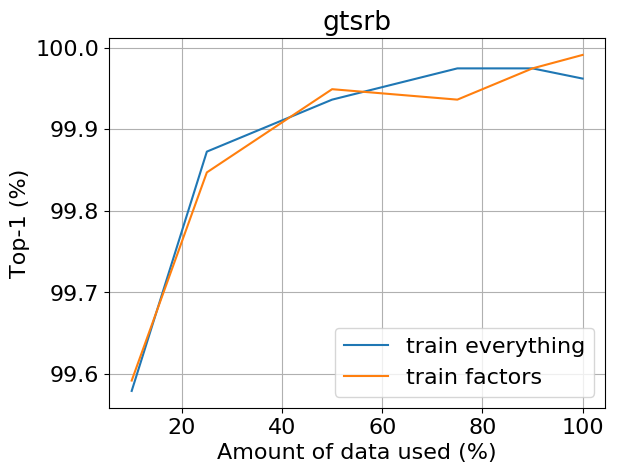} &
        \includegraphics[width=0.25\linewidth]{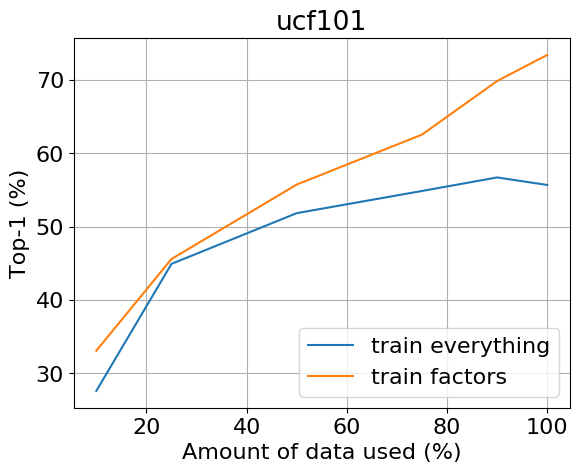} \\
    \end{tabular}
    
    \caption{Top-1 classification accuracy (\%) on  DPed, DTD, GTSRB, UFC as function of the amount of training data. Our method is compared with the performance of a network for which both the cores and the factors are fine-tuned on these datasets, also trained with the same amount of data.}
    \label{fig:partial-all}
\end{figure*}

\begin{table*}[!htbp]
	\centering
        \begin{tabular*}{0.9\linewidth}{@{\extracolsep{\fill}}  l c c c c c c c c c c }
			\toprule
			\textbf{Model}  & \textbf{Pretrained on} & \textbf{Airc.} & \textbf{C100} & \textbf{DPed} & \textbf{DTD} & \textbf{GTSR} & \textbf{Flwr} & \textbf{OGlt} & \textbf{SVHN} & \textbf{UCF}  \\
			\midrule
		     & \textbf{ImageNet}  & 55.6  & 80.7  & 99.67  & 52.2  & 99.96  & 83.8 & 88.18 & 95.66 & 78.6 \\
			\cmidrule{2-11}
			\multirow{-2.5}{*}{\textbf{Ours}}  & \textbf{Cifar100} & 41.7  & 74.5  & 99.82  & 37.55  & 99.98 & 70.9 & 88.35  & 95.43 & 72.1  \\
			\bottomrule
		\end{tabular*}
	\caption{Mean Top-1 accuracy (\%) on the unseen validation set, reported for two settings: (a) A model trained on Imagenet and adapted for the rest of the datasets, (same as the one used for the decathlon setting) (first row) and (b) a more challenging scenario where we train a model on Cifar100 and adapt it for the other datasets (second row). Notice that our method produced satisfactory results even for setting (b), marginally outperforming the Imagenet model on some datasets. This clearly illustrates the representational power of learned model and the generalization capabilities of the proposed method.}
	\label{tab:results-from-cifar100}
\end{table*}

\begin{figure}[!bh]
 \centering
 \includegraphics[width=0.8\linewidth]{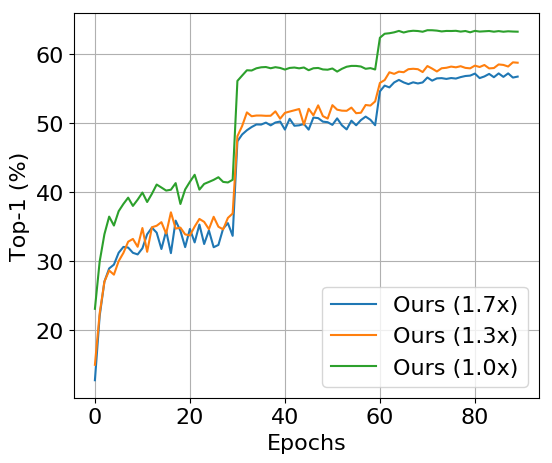}
 \caption{\textbf{Effect of the rank regularization} on ImageNet when training from scratch, for ranks achieving compression ratios of $1.0$ (full-rank), $1.3 \times$ (reducing the rank with one over the number of blocks dimension) and $1.7 \times$ (decreasing the rank of \#input and \#output channels).}
 \label{fig:compression-comparison}
\end{figure}

\begin{figure}[!h]
    \centering
    \begin{tabular}
        {
            @{\hspace{0mm}}c@{\hspace{0mm}} @{\hspace{0mm}}c@{\hspace{0mm}} @{\hspace{0mm}}c@{\hspace{0mm}} 
        }
        \includegraphics[width=0.33\linewidth]{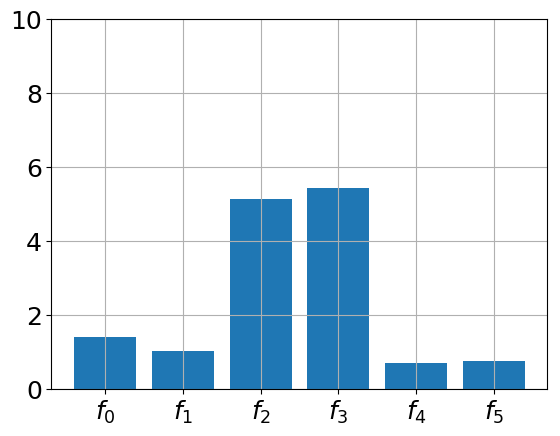} &
        \includegraphics[width=0.33\linewidth]{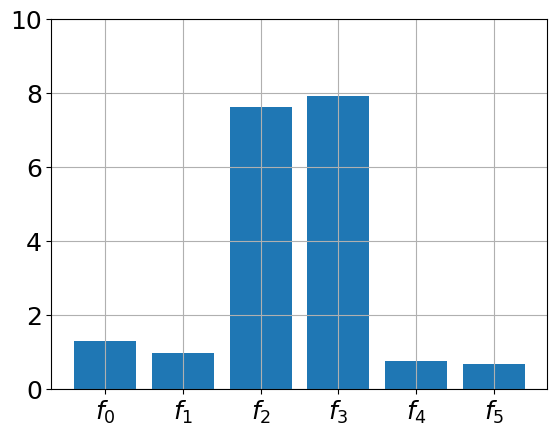} &
        \includegraphics[width=0.33\linewidth]{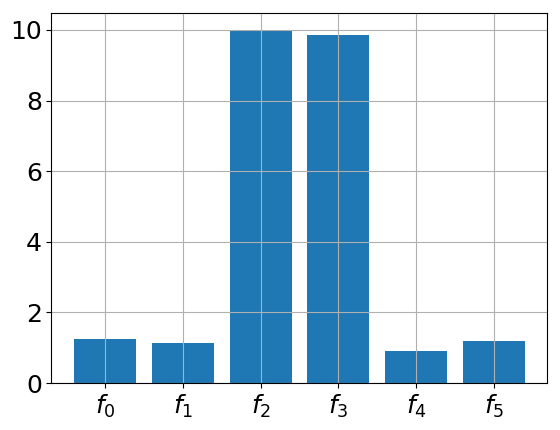} \\
        
        \includegraphics[width=0.33\linewidth]{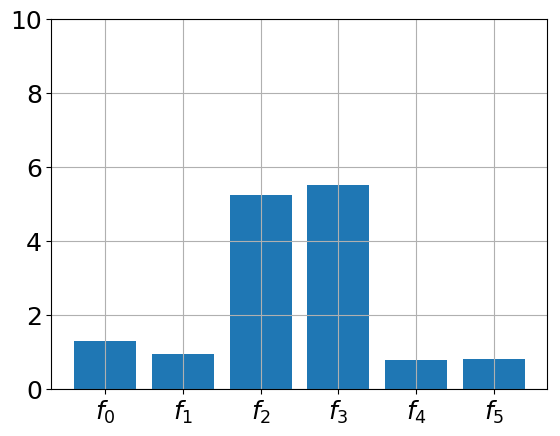} &
        \includegraphics[width=0.33\linewidth]{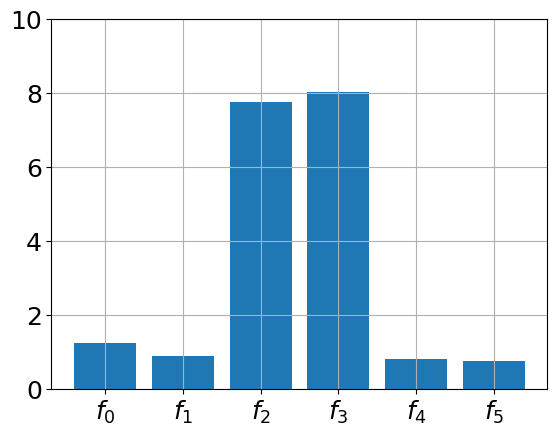} &
        \includegraphics[width=0.33\linewidth]{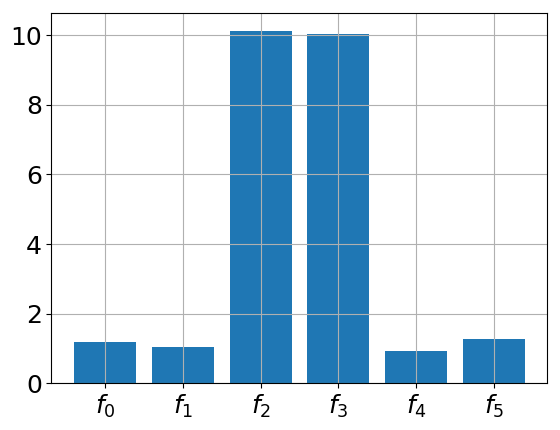} \\
        
        \includegraphics[width=0.33\linewidth]{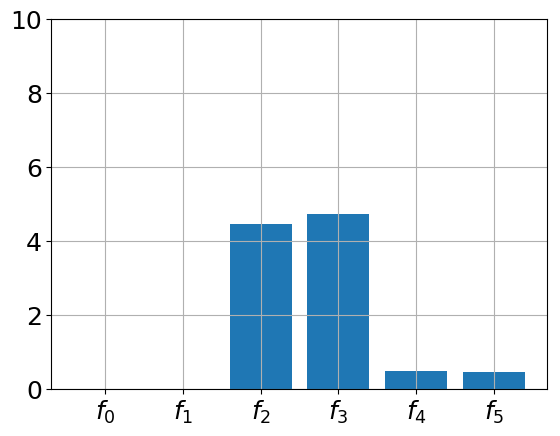} &
        \includegraphics[width=0.33\linewidth]{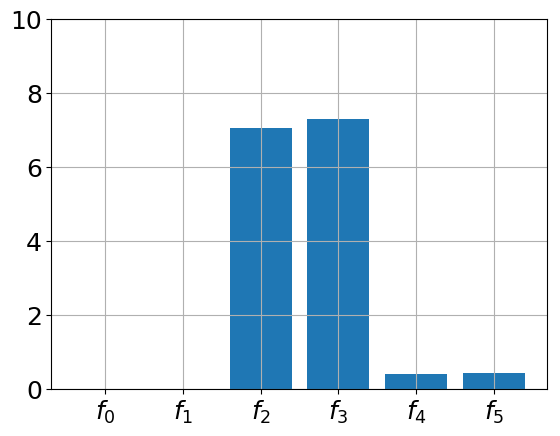} &
        \includegraphics[width=0.33\linewidth]{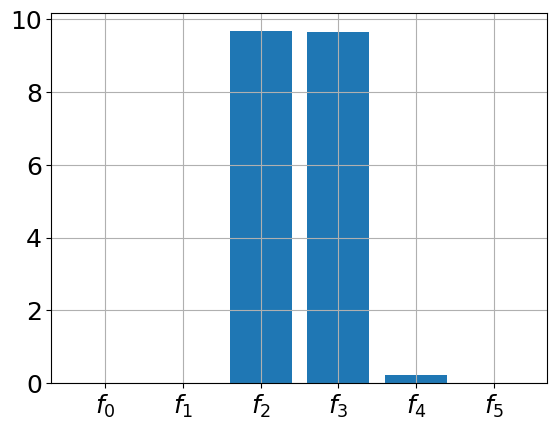} \\
    \end{tabular}

    \caption{\textbf{Orthogonality loss on the factors} on the DTD dataset.
    We show the orthogonality loss \(\mathcal{L}\) defined in Eq.~(\ref{eq:orthogonality}), for \(\lambda = 10^{-3}\) (first row), \(\lambda = 10^{-2}\) (second row) and \(\lambda = 10^{-1}\) (third row). Unsurprisingly, the orthogonality constraint is mostly violated for small values of \(\lambda\). Interestingly, we observe that the factors are almost all orthogonal, except for the dimensions \(D_2\) and \(D_3\) of the weight tensor, which correspond respectively to the number of input and output channels, confirming that these require the most adaptation from the task agnostic subspace.}
\end{figure}

\begin{table}[!h]
    \small
	\begin{center}
	   \begin{tabular*}{0.9\linewidth}{@{\extracolsep{\fill}}  l c c c c  }
			\toprule
			\textbf{Dataset}  & \bf{$\lambda=0.1$} & \bf{$\lambda=0.01$} &  \bf{$\lambda=0.001$} \\
			\midrule
		     \textbf{DTD} & 52.2\% & 51.3\% & 51.0\% \\
			 \textbf{vgg-flowers} & 80.9\% & 83.8\% & 82.2\% \\
			\bottomrule
		\end{tabular*}
	\end{center}
	\caption{Effect of enforcing weight orthogonality constraints for various values of $\lambda=\{0.1, 0.01, 0.001\}$ on two datasets: DTD and vgg-flowers. Results reported in terms of Top-1 classification accuracy (\%) on the validation set of the these datasets.}
	\label{tab:results-orthogonality}
\end{table}

\subsection{Varying the Amount of Training Data} \label{ssec:results-varying-data-amount}
An interesting aspect of incremental multi-domain learning not addressed thus far is the case where only a limited amount of labelled data available for the new domain or tasks, and how this affects performance.
Although not all 9 remaining tasks of the Decathlon assume abundance of training data, in this section, we systematically assess the sensitivity to the amount of training data. Specifically, we vary the amount of training data for 4 tasks, namely DPed, DTD, GTSRB, UFC. Fig.~\ref{fig:partial-all} shows the classification accuracy on these datasets as function of the amount of training data. In the same figure, we also report the performance of a network for which both the cores and the factors are fine-tuned on these datasets, also trained with the same amount of data. In general, especially on the smaller dataasets, we observe that our method is at least as good as the fine-tuned network which should be considered as a very strong baseline, requiring as many parameters as the original Imagenet-trained model. This validates the robustness of our model for the case of training with limited amount of training data. 

\subsection{Rank Regularization}\label{ssec:results-rank-regularization}

It is well-known that low-rank structure act as regularization mechanisms~\cite{tai2015convolutional}.
By jointly modelling the parameters of our model as a high order tensor, our model allows such constraint.
Limiting the multi-linear rank of these tensors 
effectively regularizes the whole network, thus preventing over-fitting.
This also allows for more efficient representations, by leveraging the redundancy in the multi-linear structure of the network, allowing for large compression ratios, without decrease in performance.

In this section we investigate this possibility by studying the effect of such constraint. To this end, we attempted to train our Imagenet model by imposing a low-rank constraint on the weight tensor. 
However, as can be seen  in Fig.~\ref{fig:compression-comparison}, this leads to a significant drop in performance on the base task of Imagenet; hence we did not pursue the possibility of rank regularization further. We attribute this effect to the very small number of parameters in our ResNet model.

\subsection{Effect of orthogonality regularization}\label{ssec:results-orthogonality}

To prevent degenerate solutions and facilitate learning, we added  orthogonality constraints on the task specific factors. 
This type of constraints have been shown to encourage regularization, improving the overall convergence stability and final accuracy~\cite{brock2016neural,bansal2018can}.In addition, by adding such constraints, we aim to enforce the factors of the decomposition to be full-column rank, which would ensure that the core of the decomposition preserves essential properties of the full weight tensor such as the Kruskal rank~\cite{tensor_tucker_core}. 
This orthogonality constraint was enforced using a regularization term, rather than via a hard constraint. 
See Table~\ref{tab:results-orthogonality} for results on two small datasets, namely DTD and vgg-flowers.

\section{Conclusions}
We presented a method for incremental multi-domain learning using a latent tensor factorization of the network. 
By modelling groups of identically structured blocks within a CNN as a high-order tensor, we are able to express the parameter space of a deep neural network as a (multi-linear) function of a task-agnostic subspace.
This task-agnostic core is then specialized by learning a set of small, task-specific factors for each new domain. 
While previous methods which have focused on adapting each layer separately, we show that our proposed joint modelling naturally leverages correlations across different filters and layers, resulting in a more compact representation for each new task/domain.
We evaluate the proposed method on the \(10\) datasets of the Visual Decathlon Challenge 
and show that our method offers on average about $7.5\times$ reduction in model parameters 
offering competitive results, both in terms of classification accuracy and Decathlon points.

{\small
\bibliography{AAAI-BulatA.1460.bib}
\bibliographystyle{aaai}
}

\end{document}